\documentclass{article}

\usepackage{microtype}
\usepackage{graphicx}
\usepackage{subfigure}
\usepackage{booktabs} 

\usepackage{hyperref}

\usepackage[accepted]{icml2023}

\usepackage{amsmath}
\usepackage{amssymb}
\usepackage{mathtools}
\usepackage{amsthm}

\usepackage{xspace}
\newcommand{\alg}{DIP-RL\xspace}
\newcommand{\alglong}{Demonstration-Inferred Preference Reinforcement Learning\xspace}

\newcommand{\newsec}[1]{\vspace{2mm} \noindent \textbf{#1.} }

\usepackage{color}

\usepackage[capitalize,noabbrev]{cleveref}

\theoremstyle{plain}

\theoremstyle{definition}

\theoremstyle{remark}

\usepackage{color}

\icmltitlerunning{DIP-RL: Demonstration-Inferred Preference Learning in Minecraft}

\begin{document}

\twocolumn[
\icmltitle{DIP-RL: Demonstration-Inferred Preference Learning in Minecraft}



\icmlsetsymbol{equal}{*} 

\begin{icmlauthorlist}
\icmlauthor{Ellen Novoseller}{yyy}
\icmlauthor{Vinicius G. Goecks}{yyy}
\icmlauthor{David Watkins}{sch,comp}
\icmlauthor{Josh Miller}{yyy}
\icmlauthor{Nicholas Waytowich}{yyy}

\end{icmlauthorlist}

\icmlaffiliation{yyy}{DEVCOM Army Research Laboratory}
\icmlaffiliation{comp}{Boston Dynamics AI Institute, MA, USA}
\icmlaffiliation{sch}{Columbia University, NY, USA}

\icmlcorrespondingauthor{Ellen Novoseller}{ellen.r.novoseller.civ@army.mil}
\icmlcorrespondingauthor{Nicholas Waytowich}{nicholas.r.waytowich.civ@army.mil}

\icmlkeywords{Machine Learning, ICML}

\vskip 0.3in
]



\printAffiliationsAndNotice{} 

\begin{abstract}
In machine learning for sequential decision-making, an algorithmic agent learns to interact with an environment while receiving feedback in the form of a reward signal.
However, in many unstructured real-world settings, such a reward signal is unknown and humans cannot reliably craft a reward signal that correctly captures desired behavior.
To solve tasks in such unstructured and open-ended environments, we present \alglong~(\alg), an algorithm that leverages human demonstrations in three distinct ways, including training an autoencoder, seeding reinforcement learning (RL) training batches with demonstration data, and inferring preferences over behaviors to learn a reward function to guide RL. 
We evaluate \alg in a tree-chopping task in Minecraft. Results suggest that the method can guide an RL agent to learn a reward function that reflects human preferences and that \alg performs competitively relative to baselines.
\alg~is inspired by our previous work on combining demonstrations and pairwise preferences in Minecraft, which was awarded a research prize at the 2022 NeurIPS MineRL BASALT competition, Learning from Human Feedback in Minecraft. 
Example trajectory rollouts of \alg and baselines are located at \url{https://sites.google.com/view/dip-rl}.
\end{abstract}

\section{Introduction}
\label{introduction}

In machine learning for sequential decision-making, an algorithmic agent learns to interact with an environment while receiving feedback. In particular, a typical reinforcement learning (RL) agent receives numerical reward feedback reflecting its performance; however, in many real-world settings, such a reward signal is unknown, and, furthermore, humans might not reliably handcraft a reward signal that correctly captures the desired behavior. In addition, real-world environments are often unstructured and open-ended, with complex observations and sets of possible actions, making reward shaping even more difficult. Developing algorithms to solve tasks in such open-ended and unstructured environments without rewards remains a critical challenge for artificial intelligence (AI). 

Minecraft has emerged as a state-of-the-art platform for benchmarking sequential decision-making algorithms within the machine learning research community. Minecraft shares many challenges with the real world, as it is open-ended, complex, and does not have a known numerical reward signal. In fact, the Neural Information Processing Systems Conference (NeurIPS) has recently introduced the MineRL BASALT Competition \cite{shah2021minerl}, in which competing teams train AI agents to compete in a set of four open-ended Minecraft tasks: finding a cave, building a waterfall, building an animal pen and trapping two of the same animal within it, and building a village house.




Pairwise preference-based RL has been shown to be a successful approach for learning RL reward functions when a true reward signal is unknown~\cite{christiano2017deep, lee2021pebble}. In this method, a human compares pairs of trajectory segments and indicates which behavior is preferred within each pair.
Although preference-based RL algorithms have demonstrated numerous successes, the approach inherently requires tedious human intervention and time in the decision loop. Furthermore, each pairwise preference label only provides one bit of information, potentially resulting in sample-inefficient learning. Furthermore, in early stages of learning, preference queries often consist exclusively of behavior pairs that are suboptimal or downright poor.

Demonstrations provide an effective means of acquiring examples of good behavior from the outset. Behavioral cloning (BC) is a simple and widely adopted imitation learning method that learns from demonstrations through supervised learning \cite{argall2009survey}. However, despite its popularity and ease of implementation, BC can suffer from distribution drift, which hampers its long-term efficacy. Recent state-of-the-art methods in imitation learning, such as Soft Q Imitation Learning (SQIL) \cite{reddy2020sqil}, have made progress toward addressing these issues.

Combining the strengths of imitation and preference learning, this work extends previous work on the combination of demonstrations and pairwise preferences in Minecraft \cite{shah2022retrospective} and is inspired by our solution for the 2022 MineRL BASALT Competition at NeurIPS \cite{milani2023towards}, which was awarded a research prize. This hybrid method, which we call \alglong (\alg), aims to harness the benefits of both pairwise preference learning and demonstrations by inferring preferences from an existing human demonstration dataset while avoiding the burden normally incurred by collecting pairwise preferences.

Our approach leverages the key insight that even when it is difficult for people to specify numerical reward signals or online feedback, human demonstrations may still encode significant intuition about human preference for how a task should be performed. Therefore, our approach uses human demonstrations to guide the learning process. In particular, we infer a task reward function using the demonstration data by comparing behavior segments from the expert demonstrations and from agent rollouts obtained during learning, forming a dataset of pairwise comparisons in which demonstrations are preferred to agent behaviors.
Pairwise comparison data are beneficial, since qualitative comparisons can be more reliable than handcrafted absolute numerical scores~\cite{basu2017you, sui2018advancements, joachims2017accurately}. Using pairwise comparisons between demonstrated behaviors and agent-environment interaction, we model the underlying reward that captures the desired behavior.

The contributions of this work are as follows:
\begin{enumerate}
    \item We propose \alglong (\alg), a framework for learning from human demonstrations to solve complex tasks. \alg leverages demonstrations in three distinct ways: a) to train an autoencoder that transforms images to vector embeddings, b) to seed an RL replay buffer, and c) to provide expert trajectory segments to train a reward function.
    \item We provide an evaluation of \alg in a tree-chopping task in Minecraft and compare the performance of \alg with a) human demonstrations, b) BC from demonstrations, c) RL without human demonstrations, and d) SQIL~\cite{reddy2020sqil}, a state-of-the-art imitation learning algorithm. The results suggest that \alg performs competitively relative to these baselines.
\end{enumerate}

\section{Related Work}
\label{related_work}

\subsection{Learning from Demonstrations and Preferences}

Pairwise preference-based feedback has shown promise as an intuitive method for incorporating human feedback into policy learning algorithms~\cite{10.1007/978-3-642-23780-5_11, christiano2017deep, lee2021pebble}. 
In a seminal work, \citet{christiano2017deep} take advantage of deep RL to train an agent from human feedback,
learning a deep reward model from human evaluations in the form of pairwise comparisons between trajectory segments.
The application of pairwise preferences was further advanced by active querying methods for designing informative preference queries \cite{sadigh2017active, erdem2020asking}.
PEBBLE~\cite{lee2021pebble} improves on prior preference-based RL work by leveraging unsupervised pre-training methods,
updating a policy and critic via off-policy RL, and relabeling rewards in the replay buffer as the reward model improves. These preference-based RL approaches demonstrate the ability to effectively utilize real-time human feedback to learn complex tasks.




While preference-based RL methods consider comparisons over the learning agent's behaviors, several works learn rewards from relative rankings over demonstrations~\cite{brown2019extrapolating, brown2020better, brown2020safe}. For instance, Trajectory-ranked Reward EXtrapolation (T-REX) \cite{brown2019extrapolating} employs ranked suboptimal demonstrations to infer and optimize toward a user's intent beyond the quality of the demonstrations. In contrast to works that consider pairwise comparisons purely between the learning agent's behaviors or between demonstrations, \alg leverages both demonstrations \textit{and} agent experience to generate pairwise comparisons, and thus can learn rewards while benefiting from both initial high-quality examples and from the agent's online experience.

This work extends the method presented in the 2021 MineRL BASALT competition by Team NotYourRL \cite{shah2022retrospective}.
The approach in~\citet{shah2022retrospective} builds on the method proposed in~\citet{ibarz2018reward}, and integrates Deep Q-learning from Demonstrations (DQfD) with a reward model learned from comparisons between demonstrations and agent behaviors. The team used prioritized experience replay and autolabeling of preferences to train a 
reward model. However, despite promising results, the team found that model performance was not significantly improved by this reward signal, suggesting a potential area for future research.
We propose here the use of an autoencoder that transforms images into embeddings for more sample-efficient use with RL.
Our main baseline comparison is Soft Q Imitation Learning (SQIL) \cite{reddy2020sqil}, which labels human demonstration data with +1 rewards while labeling agent experience with rewards of 0. This differs from our method in that SQIL directly labels experience with binary rewards, while \alg uses the demonstrations and agent experience to infer preferences and learn a continuous reward.

\subsection{Minecraft as a Learning Environment}

Minecraft has emerged as a popular platform for RL research due to its open-world nature and complex dynamics, and offers researchers a rich and diverse environment in which to train and test RL agents. \citet{10.5555/3061053.3061259} introduced Project Malmo, a platform for AI experimentation built on top of Minecraft that provides a sophisticated interface for RL research, opening a wide range of complex tasks to study. To promote the development of sample-efficient RL algorithms, \citet{guss2019minerl} subsequently introduced MineRL, a large-scale dataset of human demonstrations in Minecraft.

Based on the MineRL project, the MineRL Benchmark for Agents that Solve Almost-Lifelike Tasks (MineRL BASALT\footnote{MineRL BASALT Competition: \url{https://minerl.io/basalt/}.})  competition \cite{shah2021minerl} used the Minecraft environment to promote research in learning from human feedback to enable agents to accomplish tasks that lack easily-definable reward functions.
Tasks were defined by a human-readable description with no reward function.
\citet{shah2022retrospective, goecks2021combining, milani2023towards} describe some of the most promising solutions from the 2021 and 2022 competitions.

\section{Problem Setting}

We consider a learning agent that interacts with the Minecraft environment. In this setting, the agent does not observe the full-world state but instead receives an image observation based on its current location and orientation. Additionally, the agent does not observe numerical rewards. 

Therefore, we consider a reinforcement learning (RL) problem setting characterized by an episodic, partially-observed Markov decision process without rewards (POMDP\textbackslash R), $\mathcal{M} = (\mathcal{S}, \mathcal{O}, \mathcal{A}, P, P_e, \mu, T)$. Here, $\mathcal{S}$ is the underlying state space, $\mathcal{O}$ is the observation space, $\mathcal{A}$ is the action space, $P: \mathcal{S} \times \mathcal{A} \times \mathcal{S} \to [0, 1]$ yields state transition probabilities, $P_e: \mathcal{S} \times \mathcal{O} \to [0, 1]$ yields observation emission probabilities, $\mu: \mathcal{S} \to [0, 1]$ is the initial state probability distribution, and $T$ is the episode time horizon.

The agent interacts with the environment through a series of roll-out trajectories $\tau = (o_1, a_1, o_2, a_2, \ldots, o_T, a_T, o_{T + 1})$, in which the agent receives observations $o_1, \ldots, o_{T+1} \in \mathcal{O}$ and takes actions $a_1, \ldots, a_T \in \mathcal{A}$. A \textit{policy} is a mapping of observations to actions, $\pi: \mathcal{O} \times \mathcal{A} \to [0, 1]$, such that $\pi(a \mid o)$ yields the probability that the agent selects action $a \in \mathcal{A}$ given observation $o \in \mathcal{O}$.

We assume that the agent has access to a set of demonstrations of human interaction with the environment, $\mathcal{D}_{\text{demo}} = \{(o_i, a_i)\}_{i=1}^M$, where $M$ is the number of experience tuples in the demonstration dataset. 

\newsec{Learning Objective} While the environment does not include a numerical reward signal, we assume that the human demonstrations in $\mathcal{D}_{\text{demo}}$
reflect an unknown underlying reward function, $r: \mathcal{S} \times \mathcal{A} \to \mathbb{R}$. The agent's objective is to learn a behavior that maximizes $r$, such that the optimal policy $\pi^*$ is given by:
\begin{equation}
    \pi^* = \text{argmax}_{\pi} \sum_{t = 1}^T \mathbb{E}_{(s_t, a_t) \sim \mu, P, P_e, \pi} [r(s_t, a_t)].
\end{equation}

\section{\alglong}
\label{alg}

\begin{figure*}[ht]
    \vskip 0.2in
    \begin{center}
    \centerline{\includegraphics[width=2.0\columnwidth]{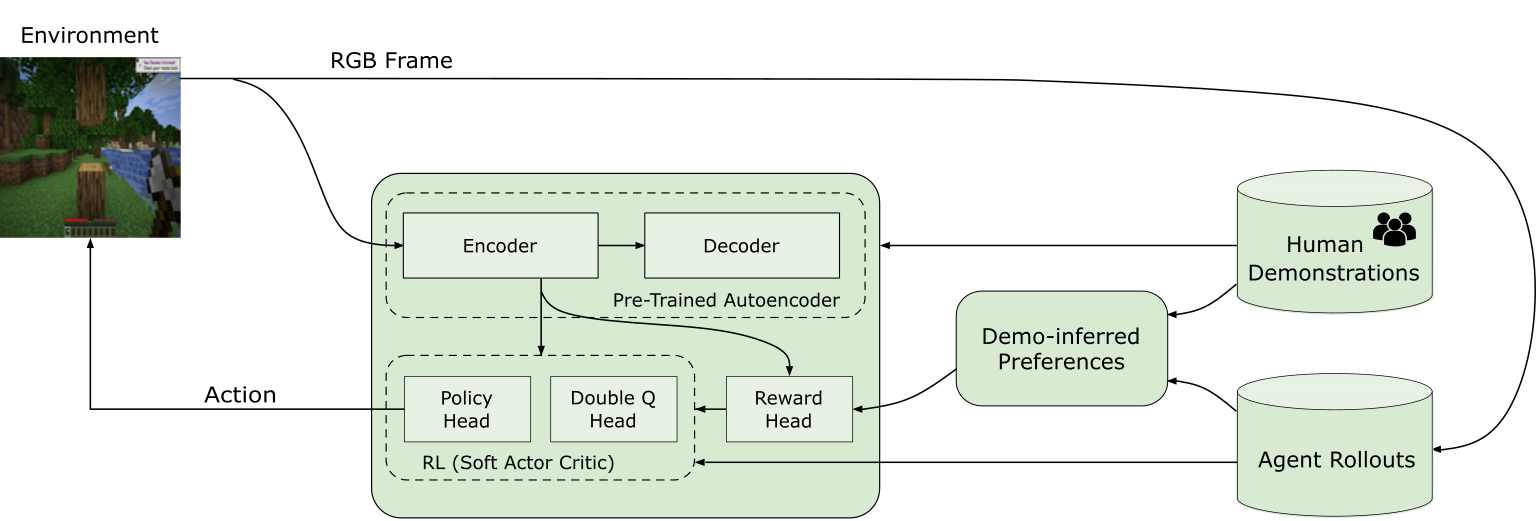}}
    \caption{System diagram of the \alglong~(\alg) algorithm. \alg~leverages human demonstrations 
    in three distinct ways: to 1) train an autoencoder to learn a compact state representation (the autoencoder training data can include nontask-specific demonstrations as well as task-specific trajectories), 2) provide trajectory segments for pairwise preference queries, and 3) provide experience to seed the RL replay buffer. The demonstration-inferred pairwise preferences are used to learn a reward function to inform a reinforcement learning algorithm.}
    \label{method_diagram}
    \end{center}
    \vskip -0.2in
\end{figure*}

We present \alglong~(\alg), illustrated in Figure \ref{method_diagram}. \alg~infers pairwise preferences from human demonstrations and agent experience
to facilitate learning. We learn a reward function from pairwise preferences that compare agent behaviors and demonstration segments. This reward function is then used to inform an RL process, similarly to preference-based RL~\cite{christiano2017deep, lee2021pebble}, but in which we also inject demonstration data. Finally, we use the demonstrations to train an autoencoder that transforms image observations into embeddings to achieve a more sample-efficient RL policy.

\subsection{Preferences between Agent and Human Behaviors}

We use pairwise preferences to learn a reward function $\hat{r}_\psi: \mathcal{O} \times \mathcal{A} \to \mathbb{R}$, parameterized by $\psi$, to inform the RL process. Each preference is given between a demonstration segment $\tau_{\rm demo}$ and an agent segment $\tau_{\rm agent}$. Although most of the work on preference-based RL compares pairs of agent behaviors~\cite{christiano2017deep, lee2021pebble}, we hypothesize that comparing demonstration-agent pairs will result in preference queries in which the demonstrated trajectory is clearly preferable; in contrast, standard preference-based RL often initially only generates preference queries involving clearly bad behaviors, since no well-performing agent behaviors are yet available.

Preference labels could be assigned either by automatically preferring demonstrated behaviors to agent behaviors or manually by a human. Our experiments consider preferences in which demonstrated behaviors are always preferred to agent behaviors, as first proposed by Team NotYourRL in~\citet{shah2022retrospective}.

Given a dataset of pairwise preferences $\mathcal{D}_{\rm pref} = \{\tau_1^{(i)} \succ \tau_2^{(i)}\}_{i=1}^N$, where $N$ is the number of pairwise preferences in the dataset, we can model the probability of each preference in terms of the learned reward $\hat{r}_\psi$ via the Bradley-Terry model~\cite{christiano2017deep, lee2021pebble}:
\begin{equation}
    P(\tau_i \succ \tau_j) = \frac{1}{1 + \exp\{-(\hat{R}_\psi(\tau_i) - \hat{R}_\psi(\tau_j))\}},
\end{equation}
where $\hat{R}_\psi(\tau) = \sum_{(o, a) \in \tau}\hat{r}_\psi(o, a)$ is the total predicted reward in trajectory $\tau$.

The reward function $\hat{r}_{\psi}$ is then optimized by minimizing the negative log-likelihood of the preference dataset $\mathcal{D}_{\rm pref}$:
\begin{equation*}
    J_{\hat{r}}(\psi) = -\log(\mathcal{D}_{\rm pref}) = -\sum_{(\tau_i \succ \tau_j) \in \mathcal{D}_{\rm pref}} \log P(\tau_i \succ \tau_j).
\end{equation*}

We regularize the reward learning objective via both a weight decay term regularizing the reward network weights and an L2-penalty on the magnitude of the predicted rewards, which we found necessary to achieve stable learning.

\subsection{Off-Policy RL via Soft Actor-Critic}

Our method leverages Soft Actor-Critic (SAC)~\cite{haarnoja2018soft} as its RL engine. Notably, \alg does not require the use of SAC, but rather could be paired with any off-policy RL algorithm. SAC is a state-of-the-art off-policy actor-critic RL algorithm that attempts to learn a policy that optimizes the maximum entropy objective:
\begin{equation}
    J(\pi) = \sum_{t = 1}^T \mathbb{E}_{(s_t, a_t)} [r_\psi(s_t, a_t) + \alpha \mathcal{H}(\pi(\cdot \mid s_t))],
\end{equation}
where $\mathcal{H}$ is entropy. SAC iterates between updating a critic (Q-function) and performing a policy improvement step. The critic is trained by minimizing the objective,
\begin{equation*}
    J_Q(\theta) = \mathbb{E}_{(o_t, a_t, o_{t+1}) \sim \mathcal{D}} \left[(Q_\theta(o_t, a_t) - \hat{Q}(o_t, a_t, o_{t+1}))^2 \right],
\end{equation*}
where $\theta$ are the parameters of the Q-network $Q_\theta$, $\mathcal{D}$ is an experience replay buffer, and
\begin{equation*}
    \hat{Q}(o_t, a_t, o_{t+1}) = \hat{r}_\psi(o_t, a_t) + \gamma \bar{V}_{\bar{\theta}}(o_{t+1}),
\end{equation*}
where $\bar{V}_{\bar{\theta}}(o)$ is the soft target value function,
\begin{equation*}
    \bar{V}_{\bar{\theta}}(o) = \mathbb{E}_{a \sim \pi_\phi(\cdot \mid o)} [Q_{\bar{\theta}}(o, a) - \hat{\alpha}\log \pi_{\phi}(a \mid o)],
\end{equation*}
where $\phi$ parameterizes the policy, $\bar{\theta}$ is a slowly-moving average of the weights $\theta$ and parameterizes the critic target network $Q_{\bar{\theta}}(o, a)$, $\hat{\alpha}$ is a hyperparameter, and the expectation is approximated via Monte Carlo estimation. As in the PEBBLE algorithm~\cite{lee2021pebble}, the most current reward model $\hat{r}_\psi$ is used to label each sampled batch of RL training data.

SAC then performs policy updates by minimizing the KL-divergence between the policy and a Boltzmann distribution given by the Q-function:
\begin{equation*}
    J_\pi(\phi) = \mathbb{E}_{o \sim \mathcal{D}} \left[KL(\pi_\phi(\cdot \mid o) \mid\mid \mathcal{Q}_\theta(o, \cdot))\right],
\end{equation*}
where $\mathcal{Q}_\theta(o, \cdot) \propto \exp\{Q_{\theta}(o, \cdot)\}$.

Finally, \alg samples RL training batches that mix together data from the experience replay buffer $\mathcal{D}$ and the demonstration dataset $\mathcal{D}_{\rm demo}$ in pre-specified proportions.

\subsection{Autoencoder and Model Architecture}

We transform image observations into vector embeddings by training an autoencoder on Minecraft image data. This approach is inspired by \citet{yarats2021improving}, in which the authors propose using an autoencoder to improve the sample efficiency of model-free RL with image observations. Similarly to \citet{yarats2021improving}, we regularize autoencoder training through both weight decay and an L2 penalty on image reconstructions. Unlike in \citet{yarats2021improving}, however, we do not continue to update the pre-trained autoencoder during RL, since we did not find that this improved performance. This may be because we pre-trained the autoencoder on a sufficiently diverse Minecraft demonstration dataset.

The image embeddings are then passed through a policy head, two Q-heads, and a reward prediction head. Note that we utilize two Q-heads, as this has shown success in previous work with model-free RL \cite{van2016deep, haarnoja2018soft}.

\section{Results}
\label{results}

\begin{figure*}[!ht]
     \centering
     \subfigure[]{
         \includegraphics[width=\columnwidth]{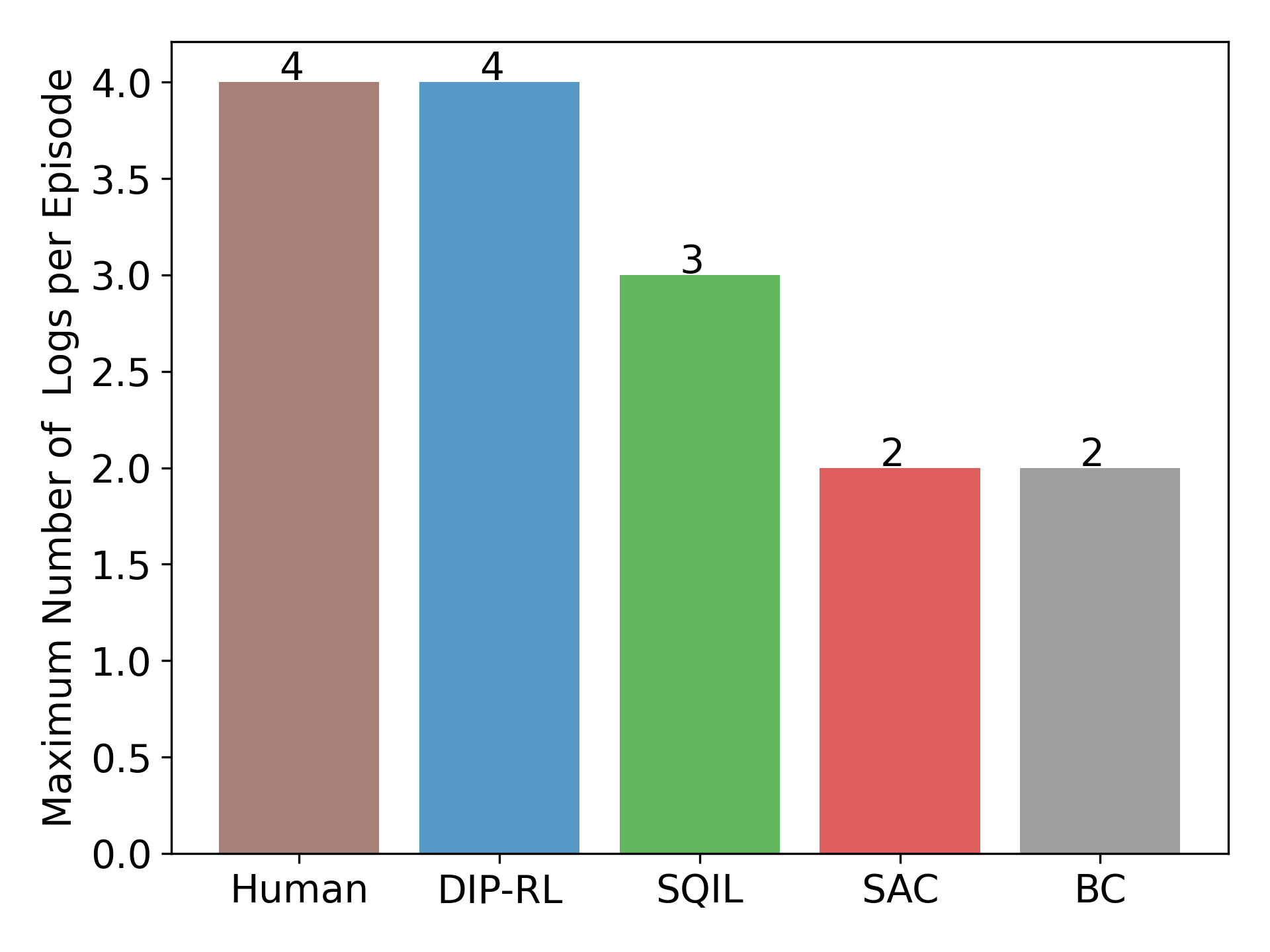}
         \label{fig:max_logs}} %
     \subfigure[]{
         \includegraphics[width=\columnwidth]{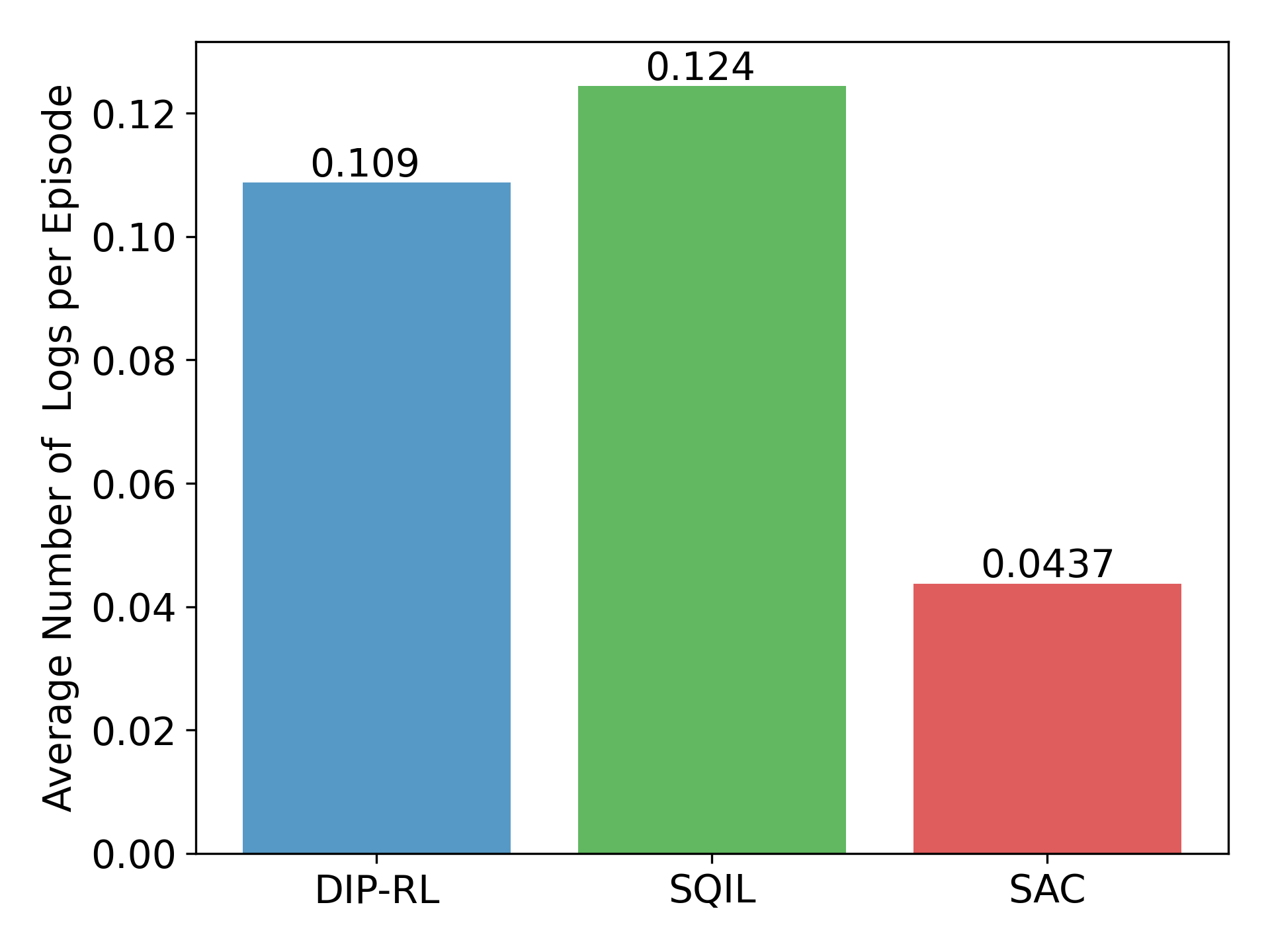}
         \label{fig:mean_logs}}
\caption{a) Maximum number of logs collected over all episodes by \alg and RL-based baseline comparisons during training, as well as by BC during evaluation (since BC does not perform environment rollouts during training) and by the human demonstrator. b) Number of logs collected per episode by \alg and each RL-based baseline, averaged over all RL training episodes.}
\label{fig:logs}
\end{figure*}

\subsection{Task Setup and Demonstration Data Collection}\label{ssec:task_setup}

We evaluate \alg and comparisons on a custom variant of the MineRLTreechop-v0~\cite{minerltreechop} environment. In this task, the agent must collect wood blocks by hitting trees in the environment.  We modify the environment to yield 128x128 image observations (rather than 64x64) and require the agent to collect a maximum of 4 logs. Because this environment provides a numerical reward signal (+1 reward every time the agent collects a log), we can straightforwardly evaluate algorithm performance; notably, however, this reward information is hidden from \alg. We fix the environment-world seed in all trials (i.e., the agent always spawns in identical surroundings), and collect 25 human demonstrations of the task.


\subsection{Methods Compared}

We compare our method, \alg, with the following baseline comparisons: Behavioral Cloning (BC) trained on the demonstration dataset, RL with SAC (which receives the numerical environment reward), and Soft-Q Imitation Learning (SQIL)~\cite{reddy2020sqil}, in which agent experience is labeled with a reward of $0$, while demonstration experience tuples are labeled with rewards of $+1$. We also report the performance achieved in the human demonstrations. 

\subsection{Performance Metrics}
In order to quantify the success of our approach, we report 1) the number of logs collected by the agent versus the number of environment steps taken during training, and 2) the maximum number of logs collected by each method compared in this work. These numbers are obtained from experience collected as part of algorithm training for the \alg, SAC, and SQIL comparisons, while BC---which does not interact with the environment during training---is evaluated after completion of training.

\subsection{Implementation Details}

We resized the Minecraft images (originally in $\mathbb{R}^{360\times640\times3}$) to RGB images in $\mathbb{R}^{128\times128\times3}$. The autoencoder was trained on the combination of 1) the TreeChop dataset described in Section~\ref{ssec:task_setup} and 2) the publicly available FindCave demonstration dataset from the BASALT competition~\cite{milani2023towards}, in which the demonstrators navigate the environment until they find a cave. We found that this combination balances task-specific images with a diverse range of Minecraft images. Each autoencoder training data batch was composed $\approx 10\%$ of images from the TreeChop demonstration data, while the remainder of the training data was drawn from the cave dataset.

Note that \alg, SQIL, and the SAC baseline all use the same pre-trained autoencoder and leverage SAC as the underlying RL method.

In the 2022 MineRL BASALT competition, the baseline agent was controlled by the hierarchical discrete action space in~\citet{baker2022video}, which comprises all possible combinations of binary buttons (e.g. attack, sprint, jump, sneak) and discretized camera commands in, by default, 11 bins each for the horizontal and vertical directions.
This scheme led to an action space too large to learn with our dataset, with 8641 possible button combinations and 121 possible camera commands. This presented an unnecessary challenge for our RL algorithms since not all button combinations are relevant for completing the proposed task.
To reduce the action space, we disabled all nonrelevant actions, giving the agent access to only the attack, move forward, move backward, and jump buttons, and restricted camera movements to a single increment to the left, right, up, and down directions.
This reduced the action space from 8641 buttons and 121 discretized camera combinations to 24 buttons and 9 discretized camera combinations.

\alg, SQIL, and SAC were each trained in an experiment run that included 800,000 steps in the environment.


\subsection{Results}


Results are illustrated in Figure \ref{fig:logs}, which presents the maximum and average numbers of logs collected by each method. For \alg, SQIL, and SAC, the maximum and average are taken over all training episodes during the algorithm run. The human demonstrations serve as an oracle baseline, since the human always collects the maximum number of logs, while BC illustrates the performance of the agent trained with no reward function, either learned or returned from the environment.

Figure~\ref{fig:max_logs} suggests that our proposed methodology, \alg, may best align with human performance in terms of the the maximum number of logs collected over all episodes. It is closely followed by SQIL, with SAC and BC demonstrating comparable performance to each other. Figure~\ref{fig:mean_logs} reflects the performance of the three RL-based methods on average over all training episodes. SQIL appears the most consistent over time, exhibiting a 13.8\% increase in the average quantity of logs collected per episode relative to our proposed \alg method.

In terms of the mean log count collected per episode, both \alg and SQIL surpass the standalone RL method, with enhancements of $149.4\%$ and $183.7\%$, respectively.

\section{Discussion}
\label{discussion}

Our results suggest that \alglong (\alg) can effectively utilize human demonstrations to learn reward functions from inferred preferences to train RL agents.
This indicates that \alg may be a valuable tool in complex and unstructured environments that lack reward signal information.

From the results presented in Figure~\ref{fig:logs}, we see that \alg is able to reach the maximum number of logs collectable in a single episode, aligning with human performance. Inconsistency in performance across training episodes suggests a possible sensitivity to initial conditions or algorithm hyperparameters.
The Soft Q-learning from Imitation (SQIL) method gathered more logs on average per episode than \alg, despite being outperformed in the maximum log count. SQIL's increased average log collection relative to \alg may occur because \alg learns from a reward signal that evolves over time, which could lead to learning instability. However, \alg may hold more potential to represent nuances in reward, since unlike SQIL, it has the potential to assign high reward labels when the agent performs well relative to the demonstrations. It would be interesting to further compare the rewards learned in \alg to the binary reward labels assigned in SQIL.

\section{Conclusion}
\label{conclusion}

This work presented \alg, an approach that uses human demonstrations in three ways to inform an RL agent. In particular, \alg uses both demonstration and agent trajectories to infer preference comparisons and learn a reward function to train RL agents in unstructured environments. We tested \alg on a tree chopping task in Minecraft and found that it could match human performance in terms of the maximum log count per episode and is competitive with baselines.

Our initial results support the potential of \alg, and of pairwise comparisons between agent and demontration behaviors, especially in scenarios where reward signals are difficult to define but demonstrations are available. In future work, we aim to improve the learning stability of \alg and to evaluate its performance across different tasks.
We believe that \alg can contribute significantly to RL and imitation learning as a method to efficiently leverage demonstrations and to infer preferences from demonstrations, and that \alg can be a useful tool for machine learning practitioners working in open-ended and unstructured environments.

\section*{Acknowledgements}

This research was sponsored by the Army Research Laboratory and was accomplished under Cooperative Agreement Numbers W911NF-23-2-0072, W911NF-18-2-0244, and W911NF-22-2-0084. The views and conclusions contained in this document are those of the authors and should not be interpreted as representing the official policies, either expressed or implied, of the Army Research Laboratory or the U.S. Government. The U.S. Government is authorized to reproduce and distribute reprints for Government purposes notwithstanding any copyright notation herein.

\bibliography{example_paper}
\bibliographystyle{icml2023}



\end{document}